\pdfoutput=1
\documentclass[conference,a4paper]{IEEEtran}
\usepackage[left=1.57cm,right=1.57cm,top=2cm,bottom=2.54cm]{geometry}
\IEEEoverridecommandlockouts

\makeatletter
\def\@IEEEauthorblockNstyle{\normalfont\fontsize{9}{10}\selectfont} 
\def\@IEEEauthorblockAstyle{\normalfont\fontsize{9}{10}\selectfont} 
\newcommand{\linebreakand}{
  \end{@IEEEauthorhalign}
  \hfill\mbox{}\par
  \mbox{}\hfill\begin{@IEEEauthorhalign}
}
\makeatother

\usepackage{cite}
\usepackage{amsmath,amssymb,amsfonts}
\usepackage{algorithmic}
\usepackage{graphicx}
\usepackage{textcomp}
\usepackage{xcolor}
\usepackage{amsmath}

\usepackage[ruled,vlined]{algorithm2e} 
\usepackage{url}
\SetNlSty{textbf}{\ }{.}  
\SetAlgoNlRelativeSize{0} %

\def\BibTeX{{\rm B\kern-.05em{\sc i\kern-.025em b}\kern-.08em
    T\kern-.1667em\lower.7ex\hbox{E}\kern-.125emX}}
\begin{document}

\title{BoMGene: Integrating Boruta–mRMR feature selection for enhanced Gene expression classification}

\author{\IEEEauthorblockN{Bich-Chung Phan}
\IEEEauthorblockA{\textit{Can Tho University}\\
Can Tho city, Vietnam \\
pbchung@ctu.edu.vn}
\and
\IEEEauthorblockN{Thanh Ma}
\IEEEauthorblockA{\textit{Can Tho University}\\
Can Tho city, Vietnam \\
mtthanh@ctu.edu.vn}
\and
\IEEEauthorblockN{Huu-Hoa Nguyen* \thanks{*Corresponding author}}
\IEEEauthorblockA{\textit{Can Tho University}\\
Can Tho city, Vietnam \\
nhhoa@ctu.edu.vn}\\
\and 
\IEEEauthorblockN{Thanh-Nghi Do}
\IEEEauthorblockA{\textit{Can Tho University}\\
Can Tho city, Vietnam \\
dtnghi@ctu.edu.vn}
}
\maketitle

\begin{abstract}
Feature selection is a crucial step in analyzing gene expression data, enhancing classification performance, and reducing computational costs for high-dimensional datasets. This paper proposes BoMGene, a hybrid feature selection method that effectively integrates two popular techniques: Boruta and Minimum Redundancy Maximum Relevance (mRMR). The method aims to optimize the feature space and enhance classification accuracy. Experiments were conducted on 25 publicly available gene expression datasets, employing widely used classifiers such as Support Vector Machine (SVM), Random Forest, XGBoost (XGB), and Gradient Boosting Machine (GBM). The results show that using the Boruta–mRMR combination cuts down the number of features chosen compared to just using mRMR, which helps to speed up training time while keeping or even improving classification accuracy compared to using individual feature selection methods. The proposed approach demonstrates clear advantages in accuracy, stability, and practical applicability for multi-class gene expression data analysis.
\end{abstract}

\begin{IEEEkeywords} Gene Expression, Feature Selection, Boruta, mRMR.
\end{IEEEkeywords}

\section{Introduction}

Gene expression data classification~\cite{b1,b2,b3} has become increasingly important in bioinformatics, effectively supporting disease subtype identification, treatment response prediction, and the discovery of potential therapeutic targets. With the rapid development of high-throughput sequencing technologies, researchers can now analyze the expression of thousands of genes simultaneously. However, this data explosion also presents major challenges in analysis and processing, often referred to as the ``curse of dimensionality''~\cite{b4}. In typical gene expression datasets, the number of genes (denoted as $p$) far exceeds the number of samples (denoted as $n$), resulting in a severe imbalance between dimensionality and sample size. As a consequence, machine learning models are prone to overfitting, with limited generalization due to many redundant or irrelevant features. Moreover, high dimensionality significantly increases computational complexity, rendering traditional classification approaches less efficient and difficult to deploy in practice.

To overcome these challenges, feature selection has been regarded as an essential step in gene expression data analysis. Identifying and retaining the most informative genes not only improves model generalization but also enhances interpretability, providing valuable biological insights. At the same time, narrowing the feature space substantially reduces computational cost and accelerates model deployment. From a biological perspective, only a small subset of genes is directly related to disease mechanisms, making feature selection indispensable in biological data analysis. Without an effective strategy, models are susceptible to noise, achieve low accuracy, and fail to identify clinically meaningful biomarkers.

Given the importance of feature selection, many studies have focused on developing robust feature selection methods tailored to gene expression data. Traditional approaches include filter-based techniques~\cite{b5,b6}, which use statistical measures such as mutual information, correlation coefficients, and entropy to rank features independently of the classifier. Although fast and computationally efficient, these methods often overlook complex interactions among genes. Wrapper-based approaches~\cite{b7,b8}, such as recursive feature elimination (RFE)~\cite{b9,b10}, refine feature subsets based on classification performance but are typically computationally expensive. Embedded methods, such as LASSO~\cite{b11,b12} or decision tree-based models~\cite{b13}, integrate feature selection into model training; however, they do not always provide optimal subsets for all classifiers. Despite significant progress, no single feature selection method consistently outperforms others across all gene expression datasets, which motivates combining the strengths of multiple techniques for better overall performance.

In recent years, gene expression classification has attracted substantial attention, yielding promising results. Feature selection, in particular, has been widely applied to reduce dimensionality, improve accuracy, and mitigate overfitting in machine learning models. ~\cite{b14} proposed a two-stage gene selection method that combines an improved mRMR with the Bat algorithm. In this approach, mRMR performs preliminary filtering of genes, and the Bat algorithm then searches for an optimal subset, markedly increasing classification accuracy on cancer microarray data. ~\cite{b15} developed ERGS (Effective Range-based Gene Selection), which evaluates the effective range of each feature via statistical analysis. This method assigns higher weights to genes with strong discriminative ability and demonstrated effectiveness on datasets related to leukemia, lung cancer, colon cancer, diffuse large B-cell lymphoma (DLBCL), and prostate cancer. In addition, several research groups have applied modern feature selection methods to discover disease-related biomarkers. For instance, one group~\cite{b17} combined Boruta and MCFS to identify 36 genes that clearly distinguish patients from healthy controls in COVID-19 diagnosis. ~\cite{b1} proposed the MC-SVM-1 model, which adopts a One-Versus-All (OVA) strategy with 1-norm SVM-based feature elimination, removing up to 99\% of redundant features and improving accuracy and interpretability compared with traditional SVM and Random Forest. Furthermore, multi-criteria hybrid approaches have received increasing attention. Zhang and colleagues proposed a two-stage mRMR-ReliefF algorithm in which ReliefF identifies informative candidate genes, and mRMR further reduces redundancy to select an optimal subset; experiments on seven different microarray datasets showed clear accuracy gains for SVM and Naive Bayes compared with using either method alone. More recently, ~\cite{b16} proposed AmRMR, an improvement over mRMR that uses Pearson correlation to quantify redundancy among features and the R-value to assess relevance to class labels. Experiments on multiple gene expression datasets showed that AmRMR improves classification accuracy over conventional mRMR, particularly for continuous-valued data. More recently, Phan et al.~\cite{b25} proposed BOLIMES, an integrated feature selection framework that combines the Boruta algorithm with LIME-based interpretability analysis to optimize gene subset selection for classification tasks. BOLIMES not only ensures the relevance and stability of selected features but also provides model-agnostic explanations for enhanced biological interpretability, demonstrating improved accuracy and interpretability across multiple gene expression datasets.

Nevertheless, despite these advances, most current studies focus primarily on selecting and ranking features rather than addressing comprehensive optimization of the feature set, including balancing relevance, redundancy, and applicability across diverse classifiers. In particular, flexible integration of each algorithm's strengths into a unified feature selection pipeline remains underexplored. To address these limitations, this study proposes \textit{BoMGene}---a compact feature selection algorithm designed to improve gene expression classification by systematically refining the feature set. Unlike traditional approaches that rely solely on statistical criteria or classifier-specific mechanisms, BoMGene combines the strengths of mRMR and the powerful elimination capability of Boruta. First, mRMR identifies and retains the most informative features while removing redundant and irrelevant ones, ensuring high relevance with minimal redundancy. Next, Boruta is applied to further reduce the dimensionality of the mRMR-selected set, thereby substantially lowering computational complexity without materially degrading classification accuracy. This combination is motivated by three reasons: (1) mRMR effectively handles global feature relevance by considering feature--feature interactions via mutual information, while Boruta performs local refinement through rigorous statistical testing. Concretely, mRMR first reduces the feature space (potentially to the order of hundreds), and Boruta then operates on this narrowed space to save training time and avoid overfitting. This two-tier approach balances dimensionality reduction with accuracy preservation, offering fast training; (2) it exploits both strategies---mRMR, as a filter method, is fast and model-agnostic for large-scale reduction, whereas Boruta, as a wrapper method, accounts for complex interactions to ensure influential variables are not overlooked; and (3) mRMR removes the bulk of redundancy, preventing Boruta from being trapped by an excessively large search space.

The main objective of this study is to obtain a compact, highly predictive feature subset. Accordingly, BoMGene offers a practical and effective solution for feature selection in gene expression data, enhancing classification performance and interpretability in bioinformatics applications. We leverage the \textit{global} strength of mRMR and the \textit{local} refinement of Boruta to retain high-quality features that capture class characteristics. This study focuses on feature selection rather than classifier design; therefore, comparisons are conducted against traditional machine learning algorithms rather than deep learning methods (e.g., ANN, KAN). 

The remainder of the paper is organized as follows: Section~II presents the theoretical foundations. Section~III details the proposed model, including its components and workflow. Section~IV reports experiments and discusses the results. Finally, Section~V concludes the paper and outlines future work.
\section{Background}
In this section, we provide formal definitions related to gene expression classification and feature selection. In addition, several classical machine-learning models are summarized to serve as baselines for comparing the effectiveness of the proposed method.

\subsection{Gene Expression Data Classification}
Gene expression data classification (GEDC)~\cite{b10,b11,b18} is an important research area in biomedicine, offering deep insights into molecular mechanisms associated with many diseases. As machine-learning (ML) methods become increasingly central in this field, they show strong potential for discovering latent patterns in datasets with large volume and high dimensionality. However, such datasets often contain many irrelevant or weakly relevant features, which introduce noise and lead to overfitting. Therefore, feature selection is crucial for retaining the most informative genes, improving accuracy and interpretability, and reducing computational cost. By focusing on key biological signals (biomarkers), recent studies aim to build reliable GEDC pipelines.

\noindent\textbf{Definition 1} \textit{(GEDC).}
Let $D=(X,y)$ be a labeled dataset, where $X\subseteq\mathbb{R}^{n}$ is the feature space and $y\in Y=\{1,2,\ldots,k\}$ is the class label. A classifier is a mapping $f:X\rightarrow Y$ that predicts $f(x)$ for an input $x$. The classifier is trained on $D=\{(x_i,y_i)\mid x_i\in X,\ y_i\in Y,\ i=1,\ldots,N\}$ by minimizing a loss function $\ell:Y\times Y\to\mathbb{R}_{+}$ that measures the discrepancy between predictions and true labels. After training, $f$ is used to classify unseen samples.

\subsection{Feature Selection Methods for Dimensionality Reduction}
Feature selection is a key preprocessing step that helps handle datasets with many features while saving computation time. This paper focuses on two techniques, Boruta~\cite{b17} and mRMR~\cite{b16,b18}, described below.

\subsubsection{The Boruta Algorithm}
Boruta~\cite{b17} is a powerful wrapper method designed to identify all truly relevant variables by comparing each original feature with randomized ``shadow'' features generated from the data. This guards against spurious relevance and preserves meaningful variables.

\noindent\textbf{Definition 2} \textit{(Feature selection with Boruta).}

Boruta selects a subset $S^\ast\subseteq X$ through the following steps:
\begin{enumerate}
    \item \textit{Shadow generation:} For each original feature $x_j\in X$, create a shadow feature $x^{\text{shadow}}_j$ by randomly permuting its values across samples, and form the augmented set $X^{\text{aug}}=X\cup X^{\text{shadow}}$.
    \item \textit{Importance estimation:} Train a classifier (typically a Random Forest) on $X^{\text{aug}}$ and compute an importance score $Z(\cdot)$ for every feature.
    \item \textit{Comparison and decision:} Let $M_{\text{shadow}}=\max_{z\in X^{\text{shadow}}} Z(z)$. For each original feature $x_j$:
    \begin{itemize}
        \item If $Z(x_j) \gg M_{\text{shadow}}$, mark $x_j$ as \emph{confirmed} (relevant);
        \item If $Z(x_j) \ll M_{\text{shadow}}$, mark $x_j$ as \emph{rejected} (irrelevant);
        \item Otherwise, mark $x_j$ as \emph{tentative}.    
    \end{itemize}
    \item \textit{Iteration:} Remove confirmed and rejected features, regenerate shadow features for the tentative set, and repeat steps 1)--3) until all features are decided or a maximum number of iterations is reached.
\end{enumerate}

\textit{Output:} $S^\ast=\{x_j:\ x_j\ \text{is confirmed}\}$.

\subsubsection{The mRMR Algorithm (Minimum Redundancy Maximum Relevance)}
mRMR~\cite{b18,b19} is an effective and widely used feature selection method in machine learning. It balances feature--label relevance with inter-feature redundancy.

\noindent \textbf{Definition 3}\textit{(Feature selection with mRMR).}
Using mutual information (MI), define the relevance of a candidate feature $X_j$ to the label $y$ as
\begin{equation}
\mathrm{Rel}(X_j,y)=I(X_j;y),
\label{eq:rel}
\end{equation}
and its redundancy with respect to the already selected set $S$ as
\begin{equation}
\mathrm{Red}(X_j,S)=\frac{1}{|S|}\sum_{X_s\in S} I(X_j;X_s).
\label{eq:red}
\end{equation}

The next feature to add is chosen by
\begin{equation}
X^\ast=\arg\max_{X_j\in X\setminus S}\big[\mathrm{Rel}(X_j,y)-\mathrm{Red}(X_j,S)\big].
\label{eq:mrmr}
\end{equation}

This process is repeated until the desired number of features is obtained or no additional feature satisfies the criterion.

In this study, BoMGene employs the FeatureWiz library---a Python tool that automates feature selection---together with mRMR to optimize dimensionality reduction.\footnote{\url{https://github.com/AutoViML/featurewiz}} This approach leverages FeatureWiz's automation and flexibility alongside the selection power of mRMR, aiming to build a compact feature set that still preserves strong representativeness for gene expression data. Specifically for the classification task, we measure feature--label relevance using the F-test and feature--feature redundancy using the Pearson correlation, rather than mutual information as in the original definition, to better suit continuous gene-expression features.

\subsection{Machine-Learning Algorithms for Gene Classification}

In this part, we examine and evaluate the performance of several leading machine-learning algorithms for gene expression classification after feature selection. First, Support Vector Machine (SVM)~\cite{b20} is used with a Gaussian RBF kernel to explore separability in high-dimensional feature spaces; it exploits margin maximization and is relatively robust to overfitting when the data are imbalanced. Next, Random Forest (RF)~\cite{b21}, an ensemble of independent decision trees, is adopted for its stability across configurations and its capability to estimate relative feature importance. In our experiments, the number of trees and maximum depth are optimized via grid search to balance accuracy and computational cost. Additionally, we include two boosting methods---eXtreme Gradient Boosting (XGB)~\cite{b22} and Gradient Boosting Machine (GBM)~\cite{b23}---to compare their strengths. XGB often achieves superior performance on large datasets thanks to its handling of missing values, learning-rate control, and built-in regularization that helps mitigate overfitting.
\section{BoMGene Algorithm}

The BoMGene algorithm provides a two-stage feature-selection solution that leverages the strengths of both mRMR and Boruta to optimize the analysis of gene-expression data. The approach prioritizes features that are highly informative with respect to class labels while eliminating redundancy and noise. First, mRMR rapidly screens a set of highly relevant, low-redundancy features to form a compact search space. Next, Boruta performs deeper screening using shadow features and Random Forest importance. This balances computational speed and statistical rigor, ensuring that the resulting set $S$ is informative and compatible with multiple classifiers. Finally, the selected set is validated by cross-validated training with SVM, RF, XGB, and GBM to identify a stable, well-generalizing subset. Algorithm~\ref{alg:bomgene} outlines the steps.

\begin{algorithm}
\caption{BoMGene: Combining mRMR and Boruta for Feature Selection}
\label{alg:bomgene}
\SetKwInput{KwIn}{Input}
\SetKwInput{KwOut}{Output}
\SetKwComment{Comment}{/* }{ */}

\KwIn{$X \in \mathbb{R}^{m\times n}$: data matrix ($m$ samples, $n$ features); $y \in \mathbb{R}^{m}$: class label vector; $T$: number of Boruta iterations}
\KwOut{$S$: optimal feature set; Classification performance on $S$} 
\Comment{B1. Initial selection with mRMR}
$S_{\text{mRMR}} \leftarrow \varnothing$; $F \leftarrow \{1,2,\ldots,n\}$;\\
\While{$F \neq \varnothing$}{
    \ForEach{$i \in F$}{
        $\mathrm{Rel}_i \leftarrow F(X_i, y)$;\\
        \If{$S_{\text{mRMR}} = \varnothing$}{
            $\mathrm{Red}_i \leftarrow 0$;
        }
        \Else{
            $\mathrm{Red}_i \leftarrow \frac{1}{|S_{\text{mRMR}}|}\sum_{j \in S_{\text{mRMR}}} |\mathrm{corr}(X_i, X_j)|$;
        }
        $\mathrm{mRMR}_i \leftarrow \mathrm{Rel}_i - \mathrm{Red}_i$;
    }
    $j^\star \leftarrow \arg\max_{i \in F} \mathrm{mRMR}_i$;
    $S_{\text{mRMR}} \leftarrow S_{\text{mRMR}} \cup \{j^\star\}$; $F \leftarrow F \setminus \{j^\star\}$;
}
$G \leftarrow X_{[:,S_{\text{mRMR}}]}$;

\Comment{B2. Refinement with Boruta}
$S_{\text{Boruta}} \leftarrow \varnothing$; Generate $G_{\text{shadow}}$ from $G$;\\
\For{$t = 1$ \textbf{to} $T$}{
    Train RF on $[G, G_{\text{shadow}}], y$;
    Compute $Z_j$ (importance) for $j \in G$ and $Z^{\text{shadow}}_k$;\\
    \ForEach{$j \in \{1, \ldots, |S_{\text{mRMR}}|\}$}{
        \If{$Z_j > \max_k Z^{\text{shadow}}_k$}{
            Mark $j$: \textit{accepted};
        }
        \ElseIf{$Z_j < \max_k Z^{\text{shadow}}_k$}{
            Mark $j$: \textit{rejected};
        }
        \Else{
            Keep $j$ for next round;
        }
    }
    $S \leftarrow \{j\ |\ j\ \text{accepted}\}$;
}
\Comment{B3. Train, evaluate classifiers}
CV $\leftarrow$ 10-fold if $m \ge 300$, else LOOCV;\\
\ForEach{model $\in$ \{SVM, RF, XGB, GBM\}}{
    Train and evaluate on $S$; Store Accuracy, Precision, Recall, F1, training time;
}
\Return{$S$, classification results}
\end{algorithm}

The algorithm takes as inputs the data matrix $X \in \mathbb{R}^{m\times n}$ with $m$ samples and $n$ features, the class-label vector $y \in \mathbb{R}^{m}$, and the number of Boruta iterations $T$. In addition, we employ cross-validation (CV): 10-fold CV when $m \ge 300$, otherwise leave-one-out. The outputs of the algorithm are the selected feature set $S$ and the performance of several classifiers (SVM, RF, XGB, GBM) trained on $S$.

\noindent The selection process consists of three main steps: \noindent\textit{(1) Initial selection with mRMR.}
In this step, mRMR is used to choose features that are highly relevant to the class label and minimally redundant with each other. Start with $S_{\mathrm{mRMR}}=\varnothing$ and $F=\{1,2,\ldots,n\}$. For each feature $i\in F$, compute the mutual information (MI) with the label to estimate relevance
{Rel}(X\_i,y) and if $S_{\mathrm{mRMR}}\neq\varnothing$, compute the redundancy with respect to the already selected set {Red}\_i
The mRMR score is then $\mathrm{mRMR}_i = \mathrm{Rel}(X_i, y) - \mathrm{Red}_i$. Select $j^\star=\arg\max_{i\in F}\mathrm{mRMR}_i$, add it to $S_{\mathrm{mRMR}}$, and remove it from $F$. Repeat until a compact subset is obtained. Let $G=X_{[:,\,S_{\mathrm{mRMR}}]}$ denote the reduced matrix.
\noindent\textit{(2) Refinement with Boruta.}
After shrinking the search space, Boruta is applied to identify the truly important features. Shadow features are created by randomly permuting each column in $G$ to form $G_{\text{shadow}}$. For each iteration $t=1,\ldots,T$, train a random forest on $[G,\,G_{\text{shadow}}]$ with labels $y$, compute the importance $Z_j$ for every original feature and $Z_k^{\text{shadow}}$ for shadow features, and compare as follows: if $Z_j>\max_k Z_k^{\text{shadow}}$, mark $j$ as \emph{accepted}; if $Z_j<\max_k Z_k^{\text{shadow}}$, mark $j$ as \emph{rejected}; otherwise keep $j$ for the next round. Finally, the accepted features form the set $S$.
\noindent\textit{(3) Train and evaluate classifiers.}
With the final set $S$, perform CV using the rule above (10-fold for $m\ge 300$, otherwise leave-one-out). Train and evaluate SVM, RF, XGB, and GBM on $S$, and record accuracy, precision, recall, F1-score, and training time for comparison.

The complexity of the algorithm is analyzed as follows. For mRMR, computing MI (or an equivalent relevance statistic) for $n$ features costs $O(m)$ each, yielding $O(m\cdot n)$. Selecting $p$ features via iterative updates of \eqref{eq:red} over the remaining candidates adds approximately $O(p\cdot +n)$, so the mRMR step costs $O(mn+pn)$.  
For Boruta, $T$ Random-Forest trainings are performed on $m$ samples and $2p$ features (original $+$ shadow). Assuming one RF training costs $O(m\,p\log m)$, the Boruta step costs $O(T\,m\,p\log m)$, with the $O(Tp)$ comparisons dominated by training.  
During evaluation, with $k$-fold CV (e.g., $k=10$), each model incurs
$O\!\big(k \times C(m,|S|)\big)$, where $C(m,|S|)$ is the training cost on $m$ samples and $|S|$ features. Overall, the complexity is $O(mn+pn)\;+\;O(T\,m\,p\log m)\;+\;O\!\big(k \times C(m,|S|)\big)$.

In the next section, we present the experimental results together with ablation studies to further justify the design choices of BoMGene.

\begin{figure*}[thbp]
\centering
\includegraphics [width=1.0\textwidth]{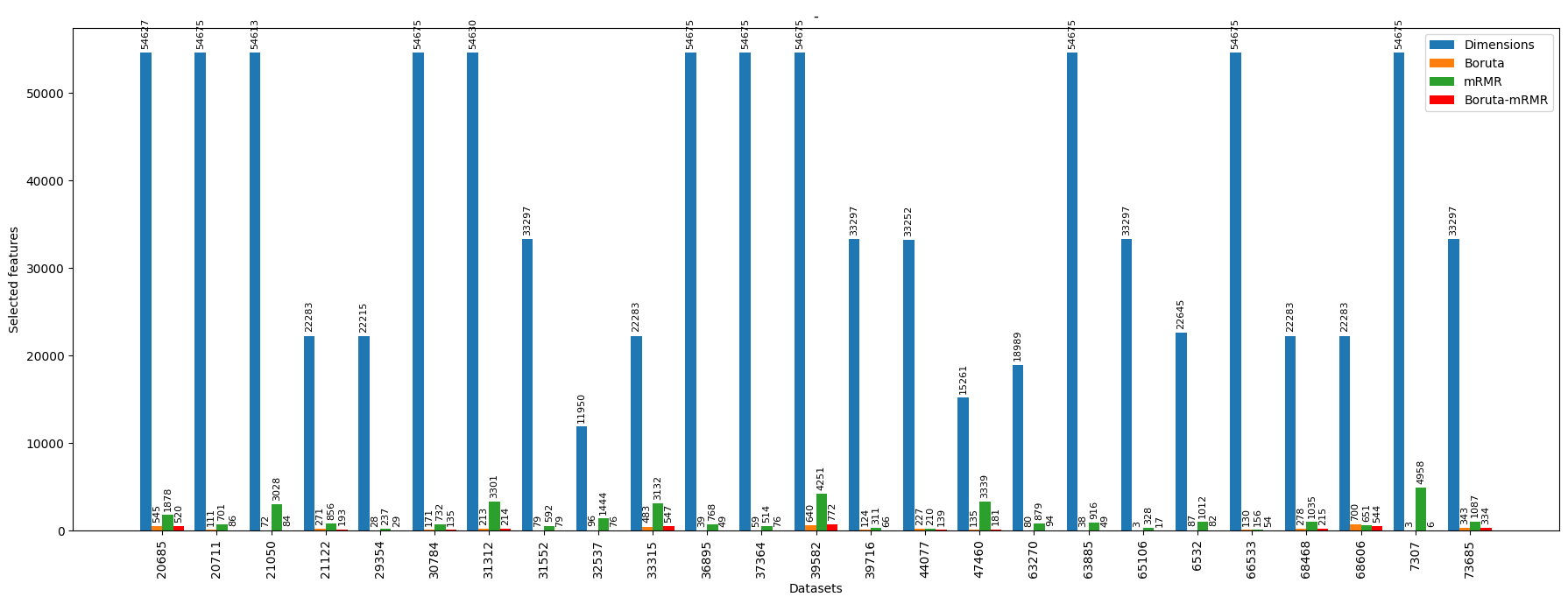}
\caption{Result of features selected by different methods }
\label{fig:FS}
\end{figure*}
\section{Experiments, Evaluation, and Discussion}

This section provides a brief description of the gene expression datasets, along with a detailed comparative analysis of the classification models. Additionally, the paper presents the results of the algorithm. The source code for this research has been made publicly available on GitHub\footnote{\url{https://github.com/PdcChung75/BoMGene}}.

\subsection{Data and Environmental Setup}

\begin{table}[htbp]
\caption{Description Of The Gene Expression Datasets}
\label{tab:datasets}
\centering
\begin{tabular}{|c|c|c|c|c|}
\hline
\textbf{ID} & \textbf{Datasets} & \textbf{Datapoints} & \textbf{Dimensions} & \textbf{Classes} \\
\hline
1 & E-GEOD-20685 & 327 & 54627 & 6 \\
2 & E-GEOD-20711 & 90  & 54675 & 5 \\
3 & E-GEOD-21050 & 310 & 54613 & 4 \\
4 & E-GEOD-21122 & 158 & 22283 & 7 \\
5 & E-GEOD-29354 & 53  & 22215 & 3 \\
6 & E-GEOD-30784 & 229 & 54675 & 3 \\
7 & E-GEOD-31312 & 498 & 54630 & 3 \\
8 & E-GEOD-31552 & 111 & 33297 & 3 \\
9 & E-GEOD-32537 & 217 & 11950 & 7 \\
10 & E-GEOD-33315 & 575 & 22283 & 10 \\
11 & E-GEOD-36895 & 76  & 54675 & 14 \\
12 & E-GEOD-37364 & 94  & 54675 & 4 \\
13 & E-GEOD-39582 & 566 & 54675 & 6 \\
14 & E-GEOD-39716 & 53  & 33297 & 3 \\
15 & E-GEOD-44077 & 226 & 33252 & 4 \\
16 & E-GEOD-47460 & 582 & 15261 & 10 \\
17 & E-GEOD-63270 & 104 & 18989 & 9 \\
18 & E-GEOD-63885 & 101 & 54675 & 4 \\
19 & E-GEOD-65106 & 59  & 33297 & 3 \\
20 & E-GEOD-6532  & 327 & 22645 & 3 \\
21 & E-GEOD-66533 & 58  & 54675 & 3 \\
22 & E-GEOD-68468 & 390 & 22283 & 6 \\
23 & E-GEOD-68606 & 274 & 22283 & 16 \\
24 & E-GEOD-7307  & 677 & 54675 & 12 \\
25 & E-GEOD-73685 & 183 & 33297 & 8 \\
\hline
\end{tabular}
\end{table}
This study experiments on 25 gene expression datasets, retrieved from the public ArrayExpress database~\cite{b24}. Detailed information about these datasets is presented in Table~\ref{tab:datasets}. Each dataset is identified by a unique code (Datasets column), primarily linked to the Gene Expression Omnibus (GEO) data source and also stored in ArrayExpress. The Datapoints column shows the number of observed samples, ranging from 53 to 677. The Dimensions column reflects the number of features, spanning from 11,950 to 54,675 dimensions, which is characteristic of the ``high-dimensional, low-sample-size'' problems commonly found in gene expression data analysis. The Classes column represents the number of classification labels, ranging from 3 to 16, reflecting the biological diversity or different pathological groupings among the research samples.

The entire process of model development and experimentation was implemented on a computer with the following configuration: Intel Core i5-12400, 2.50~GHz, 32~GB RAM, Windows 11 Pro OS, and GPU: NVIDIA GeForce RTX\texttrademark~4060 Ti. The experiments were conducted using the Python programming language with open-source libraries such as: scikit-learn (1.5.2) for traditional models and evaluation; featurewiz (0.6.1) for feature selection; pandas (2.2.3) and numpy (2.2.2) for data processing and organization; and matplotlib (3.10.3) and seaborn (0.13.2) for result visualization.

\subsection{Feature Selection Results}

After performing feature selection on the experimental datasets, the results are illustrated in Figure~\ref{fig:FS}. It can be observed that the mRMR method tends to select the largest number of features, in some cases exceeding 4,000 features. Conversely, the Boruta method selects a significantly smaller number of features, mainly ranging from a few dozen to a few hundred. Notably, the combined Boruta-mRMR method yields an intermediate result compared to the two aforementioned methods, demonstrating a certain balance between retaining informative features and eliminating redundancy. These results serve as a basis for evaluating the effectiveness of the classification models using the selected feature subsets, which will be presented in detail in the following section.

\subsection{Model Comparison and Evaluation}
To assess the effectiveness of the proposed model, we employ cross-validation together with standard classification metrics. For datasets with at least 300 samples, we use 10-fold cross-validation; otherwise, leave-one-out is applied. The average over repeated folds is reported as the performance for each dataset, which reduces dependence on a particular split and improves the reliability of the evaluation.
\subsubsection{Accuracy Evaluation} 
\begin{figure*}
\centering
\includegraphics[width=1.0\textwidth]{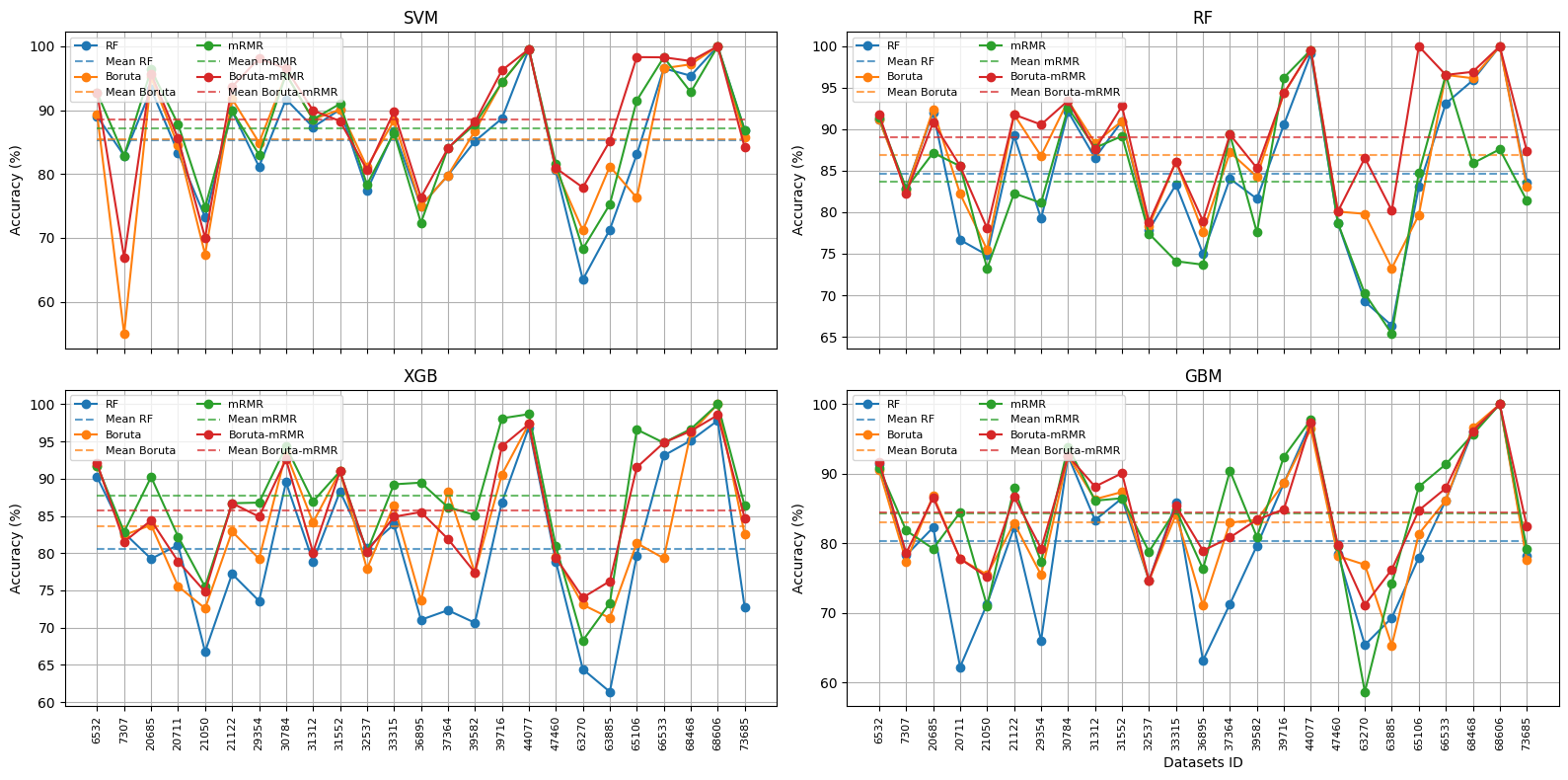}
\caption{Accuracy of the classification models}
\label{fig:acc}
\end{figure*}
Experiments on 25 gene-expression datasets reveal clear differences in classification performance across the investigated feature-selection strategies. The results are illustrated in Figure~\ref{fig:acc}, across all four classifiers (SVM, RF, XGB, and GBM), the hybrid \emph{Boruta–mRMR} method consistently attains the highest or near-highest mean accuracy compared with the baselines that use RF importance, Boruta-only, or mRMR-only (see the accuracy plots). For RF and GBM in particular, Boruta–mRMR delivers superior average accuracy while markedly reducing performance degradation on challenging datasets. Even with XGB model known to be sensitive to changes in the feature Boruta–mRMR maintains stable, top-tier accuracy, indicating that it provides high-quality and consistent features for sensitive learners. These results clearly demonstrate that combining Boruta and mRMR not only improves accuracy but also yields more stable and reliable performance in practical applications.

\subsubsection{Training-Time Evaluation}

\begin{figure*}[htbp]
\centering
\includegraphics[width=1\textwidth]{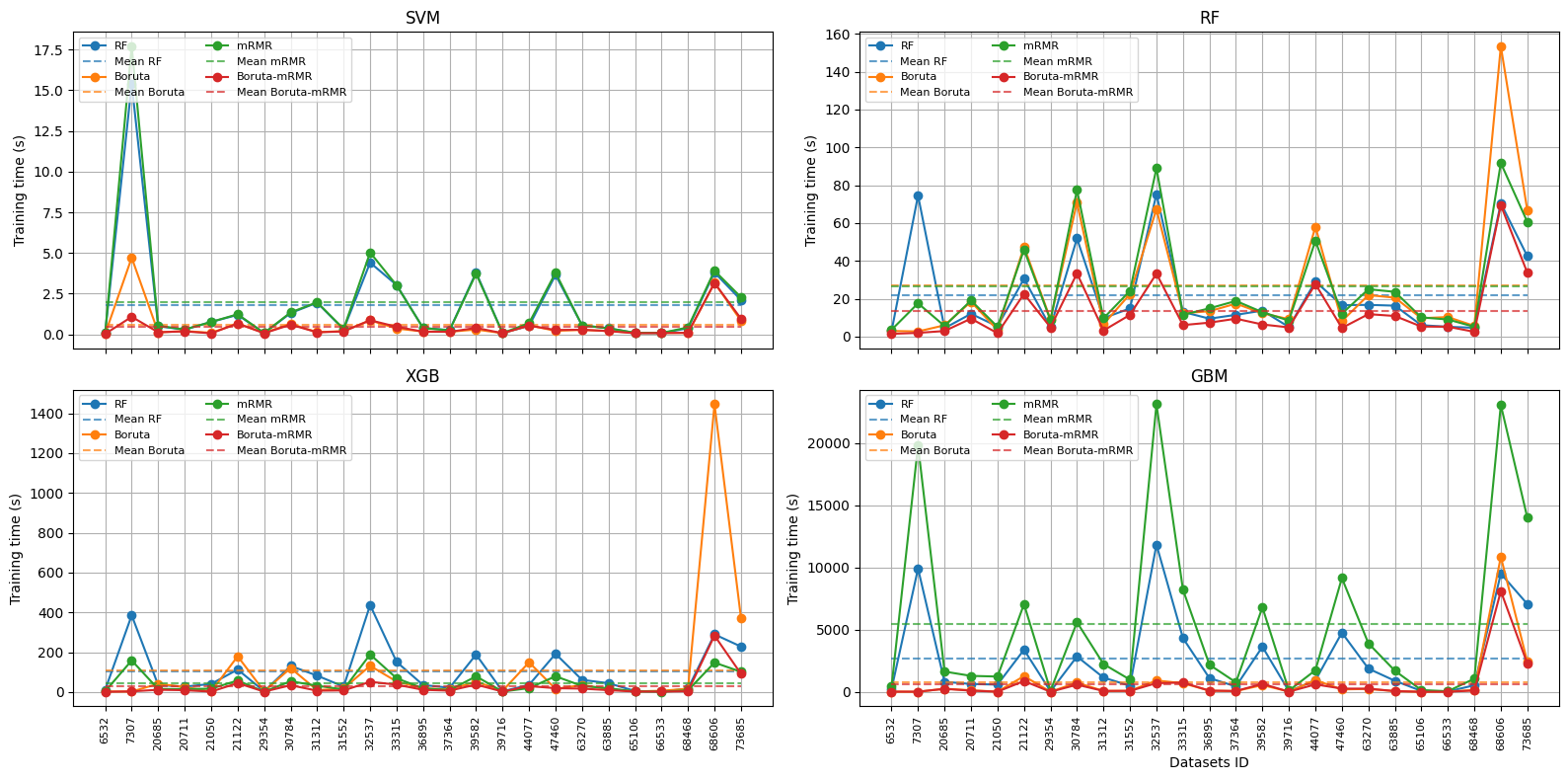}
\caption{Training time of the classification models}
\label{fig:time}
\end{figure*}

Training-time results for the feature-selection methods on the surveyed datasets with SVM, RF, XGB, and GBM (Fig.~\ref{fig:time}) show the advantage of Boruta–mRMR. Specifically, Boruta–mRMR achieves the lowest and most stable mean training time among all methods. With GBM, its mean time is 641~s, substantially lower than mRMR (5{,}456~s). With RF, Boruta–mRMR also records the smallest mean time (13~s), roughly half of RF (21~s) and less than half of Boruta (27~s). Likewise, for SVM the mean time is 0.44~s, and for XGB it is 29.55~s---both clearly lower than the alternatives. By contrast, standalone Boruta frequently exhibits large spikes, e.g., 10{,}859~s on dataset~68,606 with GBM and 153~s on dataset~68,606 with RF, indicating limitations on large/complex data. The mRMR method, while faster overall than Boruta, still shows abnormal outliers (e.g., 23{,}125~s on dataset~32,537 with GBM). RF-importance can also suffer occasional spikes (e.g., 809~s on dataset~20,685 with GBM), reflecting computational instability.

In summary, the proposed BoMGene achieves an excellent balance between feature-selection effectiveness and computational efficiency. It not only improves classification accuracy across all tested learners but also provides faster and more stable execution than traditional RF, Boruta, and mRMR baselines, confirming its practical value for real-world gene-expression classification tasks where both accuracy and time efficiency matter.

\subsection{Discussion}
Results across diverse gene-expression datasets highlight the strong balance achieved by the proposed hybrid (BoMGene - Boruta–mRMR). In terms of accuracy, BoMGene often attains the highest scores among the compared methods. For example, on dataset~29,354 it reaches \textit{98.113\%}, clearly outperforming the best baselines such as RF (81.132\%), Boruta (79.661\%), and mRMR (84.746\%). Another important advantage is the shorter and more stable training time: on dataset~21122, Boruta–mRMR requires \textit{873~s}, substantially less than mRMR (7{,}056~s) and RF (3{,}392~s), which is especially beneficial on large and complex data.

Despite these strengths, BoMGene also has limitations. In some cases, a single method may outperform the hybrid. For instance, on dataset~20711 with GBM, BoMGene attains 77.778\%, which is lower than plain mRMR (84.444\%). This can occur when the successive Boruta–mRMR filtering occasionally removes useful features due to dataset-specific properties or classifier–feature interactions. For reproducibility, the full experimental results are provided in our public GitHub repository.

\section{Conclusions and Future Work}
This study introduces a new hybrid algorithm, \textit{BoMGene}, for feature selection in gene-expression classification. Using a diverse collection of real datasets and rigorous cross-validation, BoMGene demonstrates superior accuracy compared with traditional approaches (RF, Boruta, mRMR) while also reducing training time and improving stability across all tested classifiers. 

Future work will explore improved variants of mRMR and alternative relevance/redundancy measures tailored to gene-expression data. We also plan to investigate data-augmentation strategies to mitigate overfitting and class imbalance, thereby further enhancing the overall performance of the method.

\end{document}